\documentclass[letterpaper]{article} % DO NOT CHANGE THIS
\usepackage{aaai2026} % DO NOT CHANGE THIS
\usepackage{times} % DO NOT CHANGE THIS
\usepackage{helvet} % DO NOT CHANGE THIS
\usepackage{courier} % DO NOT CHANGE THIS
\usepackage[hyphens]{url} % DO NOT CHANGE THIS
\usepackage{graphicx} % DO NOT CHANGE THIS
\usepackage{natbib} % DO NOT CHANGE THIS AND DO NOT ADD ANY OPTIONS TO IT
\usepackage{caption} % DO NOT CHANGE THIS AND DO NOT ADD ANY OPTIONS TO IT
\title{Individual Gain, Collective Loss: Metacognitive Adaptation in AI-Assisted Creativity}

\author{
    Anna Mikeda
}
\affiliations{
    Glass Umbrella\\
    anna@glassumbrella.io
}

\begin{document}
\maketitle

\begin{abstract}
Recent studies reveal a paradox: AI enhances individual creative outputs while reducing collective diversity. Current explanations---cognitive offloading and over-reliance---identify symptoms but not mechanisms. We propose selective metacognitive adaptation: routine AI use redistributes rather than uniformly diminishes metacognitive effort. Some capacities are amplified (partner modeling, surface control), while others are systematically under-supported (originality evaluation, reflective integration). This redistribution explains both individual satisfaction and collective convergence. We present a taxonomy of six metacognitive capacities organized by temporal phase, characterize their tendencies under routine AI use, and show how individually rational adaptation produces emergent social costs. The framework generates specific predictions for researchers and design principles for practitioners seeking to preserve both individual creative satisfaction and collective creative diversity.
\end{abstract}

\section{Introduction: The Creativity-Diversity Paradox}

Recent empirical studies reveal a counterintuitive pattern in human-AI creative collaboration. When individuals use generative AI tools for creative tasks, their outputs are often rated as more creative, polished, and satisfying than unassisted work. Yet when researchers examine these outputs collectively, a different picture emerges: AI-assisted creative work converges toward remarkable uniformity. Moon, Green, and Kushlev (2024) found that human-written essays contributed approximately two to eight times more to collective semantic diversity than GPT-4 essays across three studies. Doshi and Hauser (2024) documented the same pattern in creative writing: AI-assisted stories were individually rated as more creative but were significantly more similar to each other. Anderson, Shah, and Kreminski (2024) replicated this finding in ideation tasks, showing that ideas generated with ChatGPT were semantically clustered compared to human-only brainstorming.

This represents a genuine paradox: individual amplification coexisting with collective convergence. Each user experiences AI as enhancing creativity, yet the aggregate effect is a narrowing of the creative landscape. Creative diversity at the population level drives cultural innovation, scientific discovery, and the combinatorial explosion of ideas that makes collective human intelligence powerful.

\subsection{Why Current Explanations Fall Short}

Existing accounts of AI's cognitive effects tend toward broad characterizations. Terms like ``cognitive offloading'' (Risko and Gilbert 2016), ``over-reliance,'' and ``cognitive erosion'' capture something real about human-AI interaction. Yet these frameworks were not designed to explain the specific paradox of individual enhancement paired with collective convergence. They describe what happens but not why individual quality can improve while collective diversity declines.

More fundamentally, current explanations focus on what humans stop doing when AI assists them, rather than examining how human cognition adapts to the new collaborative environment. This adaptation is not passive degradation but active reconfiguration, and the pattern of that reconfiguration is what produces the paradox.

\subsection{Selective Metacognitive Adaptation}

This paper proposes selective metacognitive adaptation as the mechanism underlying the creativity-diversity paradox. Routine AI use does not uniformly diminish human cognitive engagement but redistributes metacognitive effort in systematic ways. Some metacognitive capacities are amplified through practice and strong environmental support---particularly those involved in steering AI outputs and modeling AI behavior. Other capacities are systematically under-supported, receiving neither practice nor scaffolding---especially those involved in evaluating originality and consolidating learning.

Our contribution is threefold. First, we offer a taxonomy of six metacognitive capacities relevant to creative work, organized by temporal phase and characterized by their tendency toward amplification or under-support. Second, we connect individual-level cognitive dynamics to collective-level outcomes, showing how rational individual adaptation produces emergent social costs. Third, we derive concrete empirical predictions and design implications that make the framework actionable for researchers, designers, and educators.

The tendencies we describe are based on a synthesis of existing empirical work and theoretical analysis; they represent patterns that current evidence suggests, not deterministic effects. Individual and contextual variation is substantial. This is a framework paper proposing a mechanism, not an empirical validation.

\section{Metacognition as the Critical Lens}

Metacognition---thinking about thinking---encompasses both awareness of one's cognitive processes and active regulation of those processes (Flavell 1979). In creative work, metacognition operates as the executive layer that orchestrates cognitive resources: setting goals, selecting strategies, monitoring progress, evaluating outputs, and deciding when to persist versus pivot. The distinction between ``having a creative idea'' and ``recognizing that idea as worth pursuing'' is fundamentally metacognitive. So is the difference between ``producing an output'' and ``knowing whether that output is original.''

Recent research confirms that AI-augmented creative work places substantial metacognitive demands on users. Ho{\ss}bach and Isaksen (2025) argue that collaboration with generative AI shifts part of human attention away from executing and toward monitoring and controlling creative cognition. Their analysis identifies two distinct dimensions: task-oriented metacognitive knowledge (understanding the creative problem) and AI-oriented metacognitive knowledge (understanding the AI's capabilities and limitations). Tankelevitch et al. (2024) document specific challenges: the difficulty of articulating goals clearly enough for effective prompting, the complexity of evaluating whether AI outputs meet unstated criteria, and the cognitive load of delegation decisions. Kim, So, and Park (2025) provide evidence of ``metacognitive laziness''---a tendency to prioritize rapid solution-finding over strategic planning when AI makes quick outputs available.

The cognitive offloading framework, while valuable, implies uniform transfer of cognitive work from human to external resource. But observations suggest something more structured: certain metacognitive capacities become more developed through practice with immediate feedback, while others atrophy from disuse. Users become highly skilled at prompting, iterating, and steering. They become less practiced at evaluating originality, planning exploration, and consolidating learning---capacities that current interfaces neither require nor support.

\section{A Taxonomy of Metacognitive Capacities}

\begin{table}[t]
\centering
\small
\begin{tabular}{|l|l|l|}
\hline
\textbf{Capacity} & \textbf{Phase} & \textbf{Tendency} \\
\hline
Intent formation & Before & Often bypassed \\
Exploratory planning & Before & Reduced \\
Partner modeling & During & Amplified \\
Surface control & During & Amplified \\
\shortstack[l]{Originality\\evaluation} & During/After & Under-supported \\
Reflective integration & After & Frequently skipped \\
\hline
\end{tabular}
\caption{Metacognitive capacities and their tendencies under routine AI use.}
\label{tab:metacognitive-taxonomy}
\end{table}

\subsection{Scope and Evidence Base of the Taxonomy}

The tendencies characterized in Table~\ref{tab:metacognitive-taxonomy} are synthesized from studies that vary in task type, AI tool, and participant population. The homogenization findings (Moon et al., 2024; Doshi \& Hauser, 2024; Anderson et al., 2024) draw on essay writing, short-story composition, and open-ended ideation, respectively---distinct subgenres with different evaluative norms and creative demands. Metacognitive findings (Ho{\ss}bach \& Isaksen, 2025; Kim et al., 2025; Tankelevitch et al., 2024) similarly span problem-solving, academic writing, and professional creative work.

Importantly, most tendencies in the taxonomy rest primarily on evidence from writing and ideation tasks; within-domain replication across different writing subgenres---comparing, for instance, constrained essay writing with open-ended story composition---is needed before strong subgenre claims can be drawn. Exploratory planning, for example, may be more strongly curtailed in tasks with a fixed prompt (where the first high-quality AI suggestion forecloses exploration) than in open-ended generative tasks. These differences within the writing domain are an explicit target for future empirical work, and we treat the current tendencies as directional hypotheses rather than established facts. Cross-domain validation---extending to visual design, music composition, or coding---remains a further priority.

\subsection{Before AI Interaction}

\subsubsection{Intent Formation}

Intent formation involves articulating goals, constraints, and success criteria before engaging AI. This capacity tends to be bypassed in routine use. Tankelevitch et al. (2024) identify the core tension: prompting forces writers to shift from thinking about their narrative or argument to thinking about instructions for the system. Generic framings still produce fluent outputs; idiosyncratic goals require effortful specification. The consequence is straightforward: starting with generic goals increases the likelihood of generic outputs.

\subsubsection{Exploratory Planning}

Exploratory planning involves surveying possibility space before committing to a direction. Research on human-AI creativity indicates that AI-generated suggestions can induce early convergence on initially presented solution categories, with users showing reluctance to challenge high-quality first outputs (Medeiros et al. 2025). When an AI suggestion is locally excellent, motivation to explore alternatives diminishes. This is rational individual behavior with collective costs: multiple users converging on similar first suggestions produces homogenization.

\subsection{During AI Interaction}

\subsubsection{Partner Modeling (AI Theory of Mind)}

Partner modeling involves building mental models of AI capabilities, limitations, and likely responses. This capacity is amplified through repeated interaction. Ho{\ss}bach and Isaksen (2025) identify AI-oriented metacognitive knowledge as a distinct dimension developing alongside task knowledge. Riedl and Weidmann (2025) show that Theory of Mind predicts collaborative success with AI: users with stronger ToM provide prompts that lead AI systems to engage in higher-quality dialogue. The critical nuance is that partner modeling develops in service of execution rather than evaluation. Users learn how to get AI to produce a target output, but not necessarily when AI outputs are original.

\subsubsection{Surface Control and Refinement}

Surface control involves iteratively steering outputs through editing and regeneration. This capacity is amplified and over-practiced. Current interfaces optimize for it: immediate feedback loops, easy regeneration, and visible progress. Research suggests AI tools can maintain the flow and focus of creative sessions through responsiveness (Grange et al. 2025). The concern is not that surface control lacks value, but that its amplification may mask the under-support of other capacities.

\subsection{During and After AI Interaction}

\subsubsection{Originality Evaluation}

Originality evaluation involves assessing novelty relative to others' likely work. This capacity is systematically under-supported. AI outputs are typically locally excellent: grammatically correct, logically coherent, and stylistically consistent. These surface qualities are easy to assess. Harder to evaluate is the central question: Is this original? Would someone else get essentially the same result? Current interfaces provide no scaffold for collective novelty assessment.

The empirical homogenization findings crystallize here. Moon, Green, and Kushlev (2024) found that users often did not realize their essays were converging; Anderson, Shah, and Kreminski (2024) found strong individual satisfaction despite semantic clustering. Individual assessment (``Is this creative?'') can return positive while collective assessment (``Is this different from everyone else's?'') fails---and users have little reason to perform the second assessment because the capacity to ask it is neither prompted nor supported.

\subsection{After AI Interaction}

\subsubsection{Reflective Integration}

Reflective integration involves consolidating learning from creative episodes for future work---both procedural learning (refining AI partnership strategies) and substantive integration (incorporating domain knowledge into one's own cognitive repertoire). This capacity tends to be skipped after task completion and may be the most consequential gap.

Task-focused interfaces provide no pause for reflection, and immediate output satisfaction reduces motivation to consider what was learned. Kosmyna, Tarpin-Bernard, and Rivet (2025) demonstrate the stakes: in a four-month study, participants who relied on AI assistance showed significantly reduced brain engagement and impaired memory recall compared to those who worked without AI. Critically, when AI was later removed, the AI-assisted group could not match the control group's performance---suggesting accumulated cognitive debt from episodes that could have built transferable skill but instead ended at task completion. Anderson, Shah, and Kreminski (2024) provide corroborating evidence: participants assigned significantly less responsibility to themselves for ideas generated with ChatGPT (M=48.17\%) compared to control (M=63.63\%), p=0.003, suggesting diffused ownership further reduces motivation to consolidate learning from the creative episode.

\section{Explaining the Paradox}

\subsection{Individual Level: Why Users Feel Creative}

The experience of enhanced creativity during AI collaboration is not illusory. Users genuinely exercise cognitive capacities; they simply exercise a selective subset. Partner modeling develops rapidly as users learn which prompt patterns reliably elicit desired responses. Surface control capacities strengthen through tight feedback loops that generate repeated mastery experiences.

What users often lack is not quality but distinctiveness. The capacities that would detect homogenization---originality evaluation, exploratory planning, intent formation, and reflective integration---are neither exercised nor prompted. Individual assessment (``Is this well-crafted?'') returns positive. The collective question (``Would someone else produce essentially this?'') is never asked.

\subsection{Collective Level: Why Outputs Converge}

Collective convergence emerges from two interacting mechanisms: shared starting points and absent divergence-checking. When intent formation is bypassed, users begin with generic goals that map to the same high-probability regions of model output space. When exploratory planning is curtailed by early convergence on the first AI suggestion, alternative directions go unexplored. Zhang et al. (2025) identify that AI training data creates systematic biases that limit idea diversity even when individual outputs appear novel. Meanwhile, originality evaluation---when it occurs at all---often focuses on local novelty (``Is this new to me?'') rather than collective novelty (``Is this different from what others would produce?''). The capacities required to detect convergence are precisely those that receive insufficient support.

\subsection{Why Metacognitive Adaptation Rather Than Alternative Mechanisms?}

Two alternative explanations for output convergence warrant consideration. First, convergence might simply reflect model training-data properties: AI systems trained on internet corpora naturally produce outputs that regress toward the mean of their training distribution, independent of human cognitive processes. While this is certainly true, it does not explain why humans fail to notice or compensate for this constraint. If users possessed well-developed originality-evaluation capacity, they could recognize convergence and deliberately steer away from high-probability outputs. The metacognitive adaptation account explains both the model-side constraint and why humans often do not work around it.

Second, one might argue that AI simply reduces overall task difficulty, leading to decreased cognitive engagement across the board---a simpler explanation than capacity redistribution. However, this uniform-reduction account cannot explain why some capacities (partner modeling and surface control) show clear development through practice while others atrophy. The evidence of amplified capacities contradicts the hypothesis of uniform cognitive disengagement. The pattern is differential, not uniform, which is precisely what the selective adaptation framework predicts.

Our framework is not a definitive explanation but an inference to the best explanation given current evidence. It synthesizes independent findings about metacognitive demands in AI use (Ho{\ss}bach and Isaksen 2025; Tankelevitch et al. 2024) with observed outcomes (Moon, Green, and Kushlev 2024; Anderson, Shah, and Kreminski 2024) to propose a testable mechanism connecting individual cognitive changes to collective outcomes.

\subsection{Rational Adaptation and the Social Dilemma}

The pattern described here is not cognitive laziness but rational adaptation to environmental affordances. Menary and Gillett (2022) argue that tool use involves cognitive integration rather than offloading: human cognitive capacities are genuinely transformed through tool interaction, not merely supplemented. Users of AI tools are not thinking less; they are thinking differently, in ways shaped by what their tools reward, scaffold, and ignore.

This adaptation is individually rational but aggregates into collectively suboptimal outcomes---the classic structure of a social dilemma. Creative diversity at the population level is a public good: each person's distinctive contribution enriches the pool from which others can draw. When individual users optimize for local quality without attending to collective distinctiveness, the commons of creative diversity is depleted. Awareness alone cannot resolve this; the capacities required to maintain collective diversity remain effortful and under-supported. The cumulative effect of under-supported capacities may constitute what we term cognitive debt---systematic under-development that compounds over time, though this phenomenon requires further investigation.

\section{Implications}

\subsection{For Researchers}

The framework generates specific empirical questions. Which capacities are most affected, and for whom? Validating this framework requires moving beyond self-report to behavioral traces: How long do users spend formulating prompts? Do they explore alternative directions after initial outputs? Combining such traces with collective diversity metrics would connect individual metacognitive patterns to population-level outcomes.

\subsection{For Designers}

Interface design shapes which metacognitive capacities receive environmental support. Current generative AI interfaces optimize for execution efficiency, inadvertently creating the affordance structure that produces selective adaptation. Xu, Ouyang, and Huan (2025) demonstrate empirically that metacognitive support enhances self-regulated learning in GenAI environments; Tankelevitch et al. (2024) propose a framework for metacognitive scaffolding. Capacity-targeted interventions might include structured goal articulation before AI engagement, prompts to explore alternatives before accepting initial outputs, and collective novelty indicators showing how one's output compares to typical generations.

\subsection{For Educators}

The pattern of selective adaptation suggests that metacognitive AI partnership---not just prompting skill---should be a training priority. Fan et al. (2024) provide a cautionary finding: learners using ChatGPT showed significantly higher essay-score improvement but no significant difference in knowledge gain or transfer. Kim, So, and Park (2025) demonstrate metacognitive scaffolding approaches that can mitigate these risks. Educators should consider developing reflective habits and originality-assessment skills before routine AI use, rather than attempting to retrofit them afterward.

\section{Limitations and Future Work}

This paper proposes a theoretical framework, not an empirically validated model. The tendencies we describe represent patterns that current evidence suggests, not effects we have directly measured. Individual differences in prior metacognitive development, domain expertise, and learning orientation likely moderate these patterns. The binary framing of amplified versus under-supported oversimplifies continuous, context-dependent phenomena. Our focus on routine creative work with text-based generative AI may not generalize to expert users or other domains.

Validating this framework requires studies that directly measure metacognitive capacity engagement during AI-assisted creative work. Pre-post designs comparing patterns before and after sustained AI use would test whether selective adaptation occurs as predicted. Intervention studies targeting under-supported capacities would test whether the pattern can be altered. Cross-domain validation is essential, as the homogenization findings we build on come primarily from writing tasks. Additionally, a more detailed understanding of cognitive debt---the cumulative effect of under-supported reflective integration---represents an important direction for longitudinal research.

\section{Conclusion}

We have proposed selective metacognitive adaptation as the mechanism underlying the creativity-diversity paradox in AI-assisted work. Routine AI use redistributes metacognitive effort: capacities with strong environmental support are amplified, while those without such support tend to atrophy. This redistribution explains both sides of the paradox---individual satisfaction and collective convergence---and reveals a social-dilemma structure that awareness alone cannot resolve.

From a well-being perspective, selective metacognitive adaptation is not merely a productivity or diversity concern; it is a question of cognitive flourishing. Systems that amplify execution while atrophying evaluation and reflection shape the kind of thinkers we become over time. An AI ecosystem oriented toward human well-being would be designed to strengthen the full range of metacognitive capacities, not just the subset that produces efficient outputs. This means treating the under-supported capacities---intent formation, exploratory planning, originality evaluation, and reflective integration---as design targets rather than acceptable losses. Co-evolution between human and machine intelligence, in this view, requires that machines scaffold the cognitive capacities that make humans distinctively valuable: the ability to set original goals, evaluate novelty against the breadth of collective possibility, and learn from each creative episode rather than simply completing it.

The creativity-diversity paradox is not inevitable. It emerges from a specific pattern of environmental support and neglect that current AI tools embody. Changing that pattern---through interface design that scaffolds under-supported capacities, through training that develops metacognitive AI partnership, and through collective practices that make distinctiveness visible and valuable---could preserve both individual creative satisfaction and collective creative diversity. The question is not whether AI enhances or diminishes creativity in aggregate, but which specific capacities are preserved and which are lost, and whether we design for that difference deliberately.

\section*{Appendix A. Proposed Study Design for Empirical Validation}

The selective metacognitive adaptation framework generates testable predictions. We outline a minimal study design to validate the core claims.

\textbf{Design.} Between-subjects experiment (AI-assisted vs. unassisted) with pre/post measurement, conducted across two sessions four weeks apart.

\textbf{Task.} Open-ended short essay writing on a novel prompt (minimizing prior knowledge differences).

\textbf{Operationalizations of key capacities.}
\begin{itemize}
\item \textbf{Intent Formation:} Time on task before first output (logged); structured pre-task goal-articulation prompt scored for specificity (blind rater).
\item \textbf{Exploratory Planning:} Number of distinct directions explored before committing (think-aloud protocol or branching-prompt count in AI condition).
\item \textbf{Partner Modeling:} Accuracy of predictions about AI output given a new prompt (pre-task assessment after training period).
\item \textbf{Surface Control:} Edit frequency and character-level changes per output (interaction log).
\item \textbf{Originality Evaluation:} Post-task rating of own output's novelty vs. peers' outputs (calibration accuracy against expert-panel ratings).
\item \textbf{Reflective Integration:} Recall accuracy and transfer performance four weeks later on a structurally similar task without AI.
\end{itemize}

\textbf{Primary Outcome.} Collective semantic diversity of outputs (cosine-similarity matrix, following Moon et al., 2024) and individual creativity ratings (blind expert panel).

\textbf{Prediction.} The AI-assisted group will show higher surface-control and partner-modeling scores but lower originality-evaluation accuracy and reflective-integration transfer while producing individually rated outputs of equal or higher quality with lower collective diversity.

\nocite{*}
\bibliography{ai_creativity}

\end{document}